\documentclass{article}

\usepackage[preprint]{neurips_2025}

\usepackage[utf8]{inputenc} 
\usepackage[T1]{fontenc}    
\usepackage{hyperref}       
\usepackage{url}            
\usepackage{booktabs}       
\usepackage{amsfonts}       
\usepackage{nicefrac}       
\usepackage{microtype}      
\usepackage{xcolor}         

\usepackage{graphicx}
\usepackage{subcaption}

\usepackage{ulem}
\usepackage{multirow}
\usepackage{tabularx}
\usepackage{array}  
\usepackage{adjustbox}
\usepackage{authblk}
\usepackage{amsmath}
\usepackage{amssymb}

\title{RecLLM-R1: A Two-Stage Training Paradigm with Reinforcement Learning and Chain-of-Thought}

\author{%
  Yu Xie, Xingkai Ren, YingQi, Yao Hu, Lianlei Shan  \\
  
  \texttt{zanghai1@xiaohongshu.com, lingtong1@xiaohongshu.com, yingqi1@xiaohongshu.com, huyao@xiaohongshu.com, shanlianlei@mails.ucas.edu.cn} \\
}

\begin{document}

\maketitle

\begin{abstract}
  Traditional recommendation systems often grapple with "filter bubbles", underutilization of external knowledge, and a disconnect between model optimization and business policy iteration. To address these limitations, this paper introduces RecLLM-R1, a novel recommendation framework leveraging Large Language Models (LLMs) and drawing inspiration from the DeepSeek R1 methodology. The framework initiates by transforming user profiles, historical interactions, and multi-faceted item attributes into LLM-interpretable natural language prompts through a carefully engineered data construction process. Subsequently, a two-stage training paradigm is employed: the initial stage involves Supervised Fine-Tuning (SFT) to imbue the LLM with fundamental recommendation capabilities. The subsequent stage utilizes Group Relative Policy Optimization (GRPO), a reinforcement learning technique, augmented with a Chain-of-Thought (CoT) mechanism. This stage guides the model through multi-step reasoning and holistic decision-making via a flexibly defined reward function, aiming to concurrently optimize recommendation accuracy, diversity, and other bespoke business objectives. Empirical evaluations on a real-world user behavior dataset from a large-scale social media platform demonstrate that RecLLM-R1 significantly surpasses existing baseline methods across a spectrum of evaluation metrics, including accuracy, diversity, and novelty. It effectively mitigates the filter bubble effect and presents a promising avenue for the integrated optimization of recommendation models and policies under intricate business goals.

\end{abstract}

\section{Introduction}

In the digital age, recommender systems have become indispensable tools for user navigation of vast information landscapes. However, traditional methodologies, primarily leveraging historical interaction data and techniques such as collaborative filtering and matrix factorization, exhibit inherent shortcomings. Although effective in data-rich environments for capturing certain facets of user preference, these approaches often overlook valuable external knowledge, thereby contributing to the well-established "filter bubble" effect ~\cite{koren2009matrix}. Moreover, industrial deployments frequently maintain a separation between model optimization and strategic iteration. This division, often manifested through distinct teams dedicated to model tuning versus strategy adjustment ~\cite{covington2016deep, liu2017cascade, qin2022rankflow, luo2024integrating}, can impede the rapid adaptation necessary to address evolving business requirements.

Recent advancements in large-scale pre-trained language models (LLMs) have introduced a transformative opportunity for the recommender system domain. These models demonstrate remarkable capabilities in semantic understanding and knowledge reasoning, and are capable of learning new tasks with limited data ~\cite{brown2020language, devlin2019bert, luo2024integrating, li2023gpt4rec, yang2023palr, liao2024llara, ji2024genrec}. Capitalizing on this potential to reshape data construction and model architecture in recommender systems holds the promise of overcoming the inherent limitations of traditional approaches, leading to enhancements in both the accuracy and diversity of personalized recommendations. Concurrently, it opens novel avenues for the flexible customization of business logic and strategy modules.

The current mainstream methodologies in recommender systems can be broadly categorized as follows, \textbf{Deep Learning-based Methods}: Approaches such as neural collaborative filtering and traditional collaborative filtering algorithms heavily rely on extensive sequences of user behavior data. While these methods can effectively capture certain user preferences given sufficient data, they often struggle to incorporate external semantic information. Consequently, they are susceptible to becoming trapped in local optima and face challenges in mitigating the filter bubble effect ~\cite{he2017neural, cheng2016wide, zhou2018deep}.
\textbf{Leveraging LLMs to Augment Traditional Recommender Systems}: These methods employ pre-trained language models to extract user and item features, which are subsequently fed into traditional recommender models for further optimization. Although this approach can offer some improvement in recommendation performance, it remains constrained by the architecture and training paradigms of traditional models, thus limiting the full exploitation of large-scale language model capabilities.
\textbf{Directly Employing LLMs for Recommendation}: In this paradigm, historical user behaviors and item information are directly provided as prompts to a large language model, which then generates a recommendation list. While conceptually simple, this method often proves inadequate for addressing the diverse business requirements encountered in real-world applications, still necessitating dedicated strategy modules to support specific business objectives ~\cite{rajput2023recommender}.

To address the aforementioned limitations, we propose a novel recommender system framework inspired by the DeepSeek R1 architecture ~\cite{guo2025deepseek}, with enhancements in the following two key aspects:
  \paragraph{Data Construction Optimized for Large Language Models.}
User-side: We integrate user profiles with historical behavior information, constructing prompt inputs that combine static user attributes and dynamic interaction data. This ensures that the large model can capture both the basic characteristics and behavioral habits of the user.
Item-side: We comprehensively leverage item descriptions, titles, and content-derived tags to provide a multi-faceted semantic representation of the items. Furthermore, we generate positive samples by constructing an item sequence based on the user’s subsequent interactions, guided by these tags.
Label Construction: We generate the training label, an item sequence that accurately reflects the user's preferences, based on their subsequent behavior data. This provides a more precise supervisory signal for model training.
  \paragraph{Two-Stage Training Strategy.}
SFT Stage: In the Supervised Fine-Tuning (SFT) stage, we employ a distilled version of DeepSeek R1 to fine-tune the large language model using the constructed prompt data derived from recommender system user interactions. This process activates the model’s inherent capabilities for recommendation tasks, facilitating knowledge transfer and initial model adaptation.
GRPO Stage: Building upon the SFT stage, we introduce a GRPO training strategy based on reinforcement learning. By incorporating the Chain-of-Thought (CoT) mechanism, the large language model is endowed with multi-step reasoning and holistic decision-making capabilities during the recommendation process ~\cite{shao2024deepseekmath}. Simultaneously, by flexibly defining the reward function, the model receives positive reinforcement for recommendation accuracy (e.g., click-through rate and conversion rate) and gains a parameterized customization space for diverse business scenarios. This strategy even holds the potential to supplant the diversity strategy modules and complex strategy weighting mechanisms prevalent in traditional recommender systems.

Overall, the proposed method not only mitigates the "filter bubble" effect inherent in traditional recommender models but also fully exploits the rich external knowledge embedded in large language models and the multi-step decision optimization capabilities of reinforcement learning. This culminates in a more adaptable and comprehensively optimized recommendation solution for real-world business environments.

\begin{figure}
    \centering
    \includegraphics[width=0.7 \linewidth]{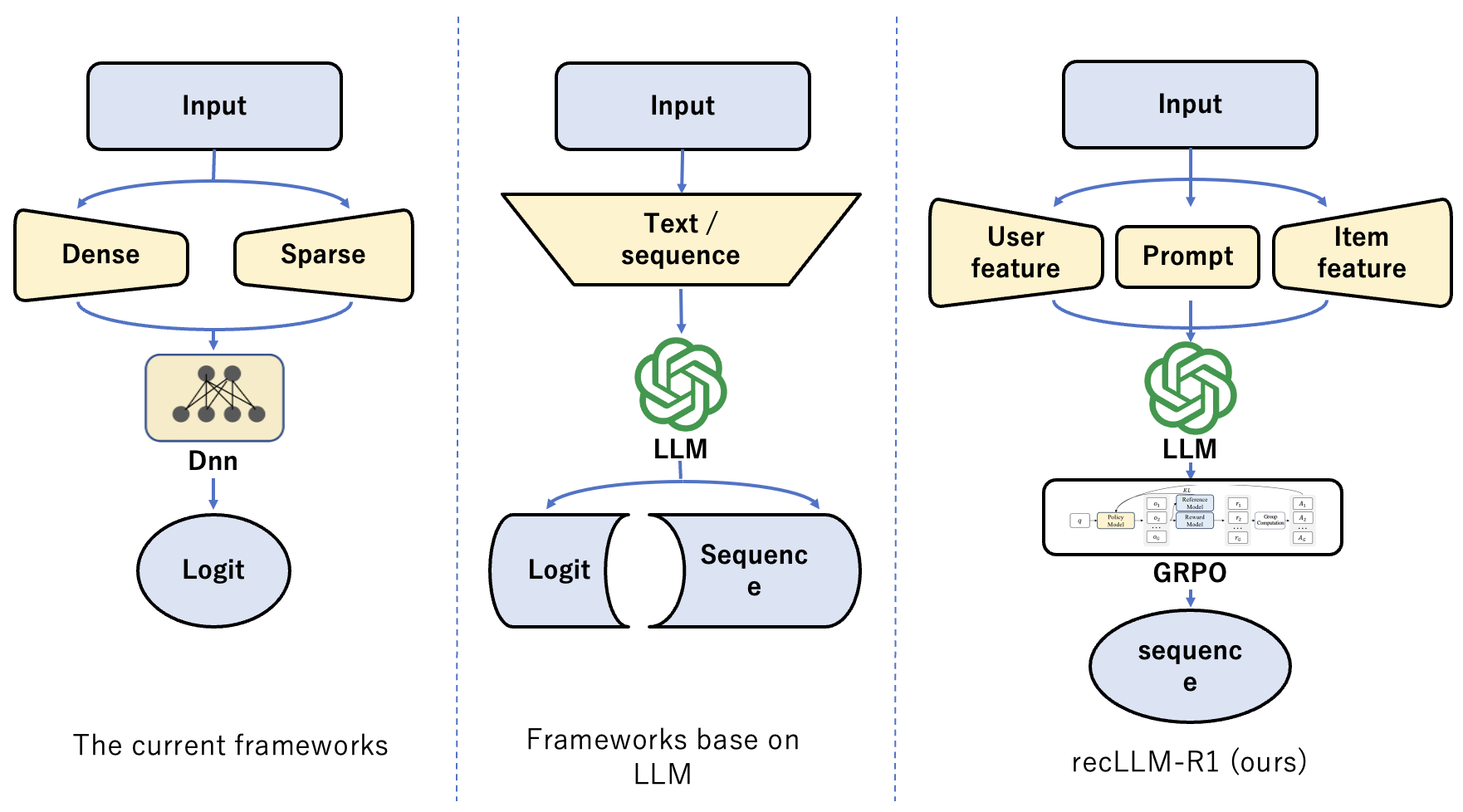}
    \caption{Comparison of three mainstream recommendation frameworks, Current frameworks / Frameworks base on LLM / recLLM-R1(ours)}
    \label{fig: frameworks}
\end{figure}

\begin{figure}
    \centering
    \includegraphics[width=0.9 \linewidth]{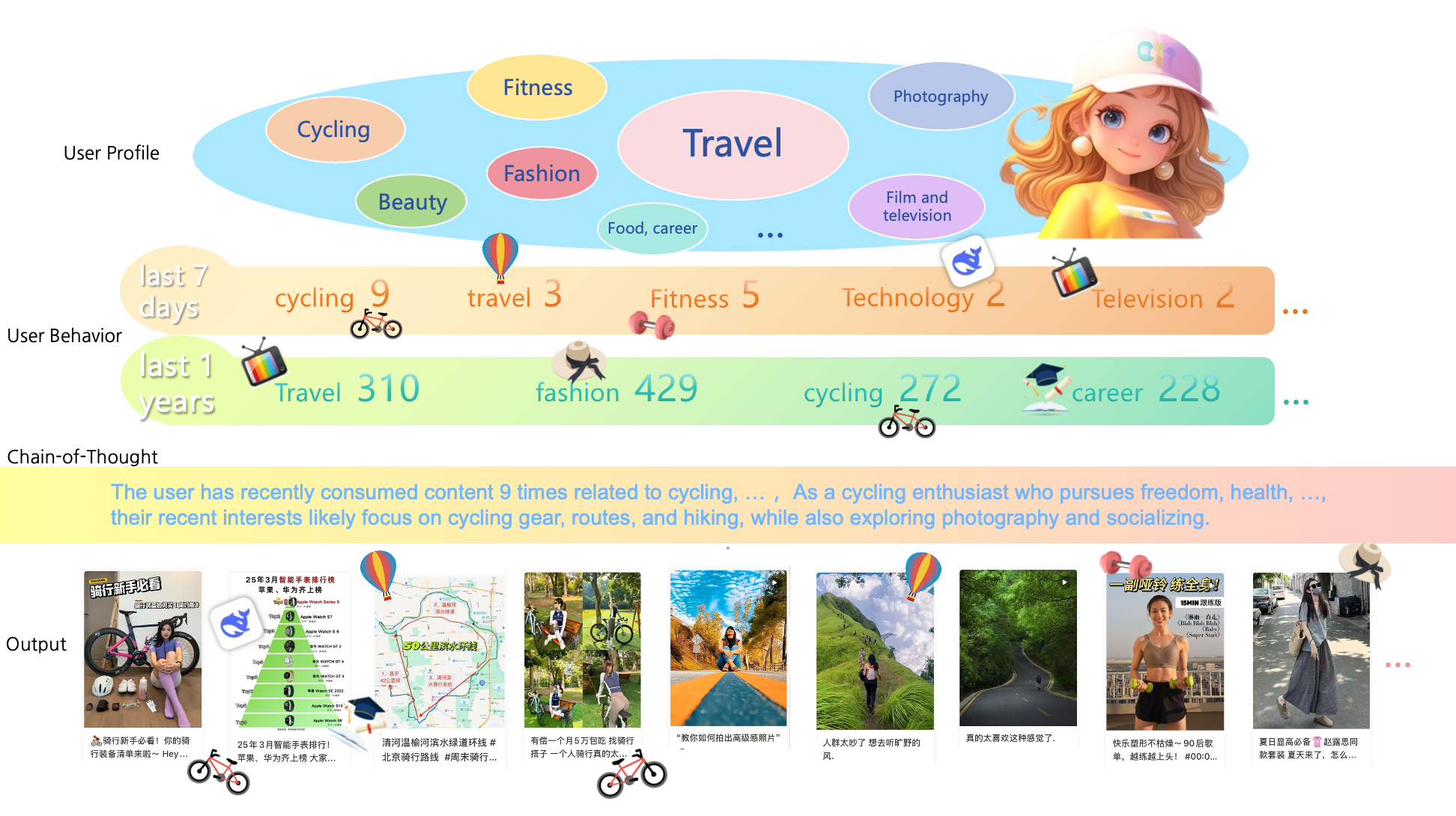}
    \caption{Illustrates the motivation behind our research}
    \label{fig: result-pic}
\end{figure}

The primary contributions of this study can be summarized as follows:
\begin{itemize}

\item \textbf{Optimized Prompt-Oriented Data Construction for LLMs.} We propose a data construction method optimized for large language models that effectively integrates user profiles, historical behaviors, and multi-modal item information to generate high-quality prompt data that meets the requirements of complex recommendation scenarios.

\item \textbf{Novel Recommendation Framework.} We propose RecLLM-R1, an innovative LLM-based recommendation framework illustrated in Figure ~\ref{fig: frameworks}. To the best of our knowledge, this is the first work to combine a two-stage SFT+GRPO training paradigm with Chain-of-Thought (CoT) reasoning specifically tailored for recommendation tasks. Drawing inspiration from the DeepSeek R1 architecture ~\cite{guo2025deepseek}, our training framework first activates the large language model’s capabilities through an SFT stage, followed by a GRPO stage incorporating reinforcement learning and the CoT mechanism. This effectively enhances the model’s comprehensive performance in long-sequence decision-making.

\item \textbf{Flexible Reward Function Design.} The reward function within our framework extends beyond traditional recommendation evaluation metrics to accommodate diverse business requirements, offering the potential to supplant dedicated diversity strategy modules in existing recommender systems. This enhances overall flexibility and adaptability.

\item \textbf{Real-world Data Validation and Open Sourcing.} We validate our model using real user behavior data from a large-scale social media platform, recognized for its rich user-generated content. Empirical results demonstrate the superior performance of our proposed model across key evaluation metrics, including recommendation accuracy, information diversity, and business flexibility. Furthermore, all code will be open-sourced and is included in the supplementary materials, providing robust support for collaboration between industry and academia.

\end{itemize}

Overall, Figure~\ref{fig: result-pic} illustrates the motivation behind our research. Our approach offers a novel perspective on recommender system design by seamlessly integrating the strengths of large language models and reinforcement learning. It effectively addresses the limitations of traditional methods—particularly the issue of information silos—while providing a solid technical foundation for multi-objective and customized recommendation in real-world business scenarios.

\section{Related Work}
\label{sec: Related Work}

\subsection{Reinforcement Learning and Chain-of-Thought}

Reinforcement Learning (RL) has increasingly attracted attention in recommender systems in recent years ~\cite{afsar2022reinforcement,ouyang2022training,stiennon2020learning,rafailov2023direct, jeong2023factual}. Its core idea is to design a reward function that enables the model to achieve long-term optimality through continuous, multi-step decision-making. Early work—such as the Deep Q-Network (DQN) proposed by Mnih et al. ~\cite{mnih2015human}, demonstrated the potential of deep RL in Atari games. However, conventional RL methods typically rely on single-step updates, making it difficult to capture the complex multi-step reasoning involved in the decision-making process. This limitation becomes apparent when dealing with the dynamic changes in user behavior and long-term dependencies.

To address this issue, the Chain-of-Thought (CoT) approach was introduced to endow models with the ability to reason step by step. The CoT method encourages large models to generate a series of intermediate reasoning steps before arriving at a final answer, thereby enhancing the transparency and interpretability of their decisions  ~\cite{wei2022chain}. For example, in a recommendation scenario, it can be explained as "the user clicked on A because A is similar to B, which the user previously clicked on." This multi-step reasoning mechanism is expected to be combined with reinforcement learning strategies by designing tailored reward functions that encourage the model to generate coherent and logical reasoning paths, thus better supporting global decision-making and meeting business requirements. Furthermore, some RL algorithms, such as Proximal Policy Optimization (PPO), have shown promising performance in these continuous decision-making tasks  ~\cite{schulman2017proximal}.

\subsection{Large Language Models}

In recent years, the emergence of large-scale pre-trained language models (LLMs), such as GPT-3 , GPT-2, and the subsequent GPT-4, along with open-source models like LLaMA and DeepSeek R1  ~\cite{brown2020language,guo2025deepseek,touvron2023llama,achiam2023gpt}, has brought revolutionary progress to natural language understanding and knowledge reasoning. These models possess powerful text generation and few-shot learning capabilities, allowing them to deeply analyze the latent semantics and sentiment tendencies present in user queries, reviews, and product descriptions. Moreover, by utilizing long-sequence modeling techniques  ~\cite{dao2022flashattention}, these models can capture fine-grained correlations in users' historical behaviors, thereby laying the foundation for personalized recommendations.

In recent years, techniques based on prompt engineering have also emerged. Such techniques can activate large models to perform diverse tasks with only a small number of prompts, offering a new paradigm for recommender systems. Combined with reinforcement learning strategies—like PPO, which can further adjust the recommendation strategy to optimize for long-term user satisfaction ~\cite{schulman2017proximal} — these large models are able to harness extensive external knowledge while continuously optimizing decisions through dynamic feedback.

DeepSeek R1 Training Process: Notably, the training process of DeepSeek R1 employs Supervised Fine-Tuning (SFT) for "cold start," followed by optimization using Group Relative Policy Optimization (GRPO) ~\cite{guo2025deepseek}. In some cases, multiple iterations of SFT and Reinforcement Learning (RL) are conducted. Conversely, R1-Zero explores the possibility of initiating training directly from the base model using pure GRPO. Our RecLLM-R1 framework is directly inspired by this SFT+GRPO structure.

GRPO Algorithm ~\cite{shao2024deepseekmath}: The GRPO algorithm optimizes based on group-wise relative rewards and does not require complex critic networks to estimate state-value functions. Considering that recommendation tasks often involve complex and multidimensional objectives (such as accuracy, diversity, novelty, and business metrics), designing a precise scalar reward function or training an accurate critic network for these complex goals is extremely challenging. GRPO relies on relative ranking of a batch of generated recommendation sequences (which may include Chain-of-Thought, or CoT), making it well-suited for optimizing complex multi-objective reward functions. Determining whether Recommendation A (and its CoT) is superior to Recommendation B under certain complex criteria is generally easier than assigning an exact absolute value score to A or B. Therefore, compared to methods like PPO, which require a critic to estimate complex values, or DPO, which typically relies on simpler pairwise preferences, GRPO may be inherently more suitable for handling the complex, multi-objective optimization problems required by modern recommendation systems. The choice of GRPO is driven not only by efficiency considerations but also by its better alignment with the complex business logic in the industry.

\subsection{Recommender Systems}

Recommender systems, as a critical technology for information filtering and personalized services, have evolved from traditional collaborative filtering to deep learning-driven multimodal fusion approaches. Early methods primarily relied on collaborative filtering and matrix factorization. For example, Koren  ~\cite{koren2009matrix} used matrix factorization to extract latent relationships between users and items. Although these methods are simple and efficient, they face significant limitations when addressing issues such as data sparsity, cold start, and the underutilization of external knowledge.

With the advent of the deep learning era, Neural Collaborative Filtering (NCF) methods have been proposed, which leverage multi-layer perceptrons to model users and items non-linearly, thereby capturing latent features more effectively ~\cite{cheng2016wide}. In addition, Transformer-based sequential recommendation models, such as SASRec  ~\cite{kang2018self} utilize self-attention mechanisms to effectively model the temporal dependencies in users’ behavior sequences.

Although these methods have improved recommendation performance to some extent, they generally face the following challenges: they tend to ignore the incorporation of external world knowledge, thus easily falling into filter bubbles; furthermore, in industrial settings, due to business needs, companies cannot solely focus on model optimization—there is often heavy reliance on business strategy online, leading to poor maintainability. Therefore, solely optimizing the model side may not be sufficient to address such issues  ~\cite{peng2024large, wang2020kerl,chang2023twin,guo2017deepfm}.

\begin{figure}
    \centering
    \includegraphics[width=0.85\linewidth]{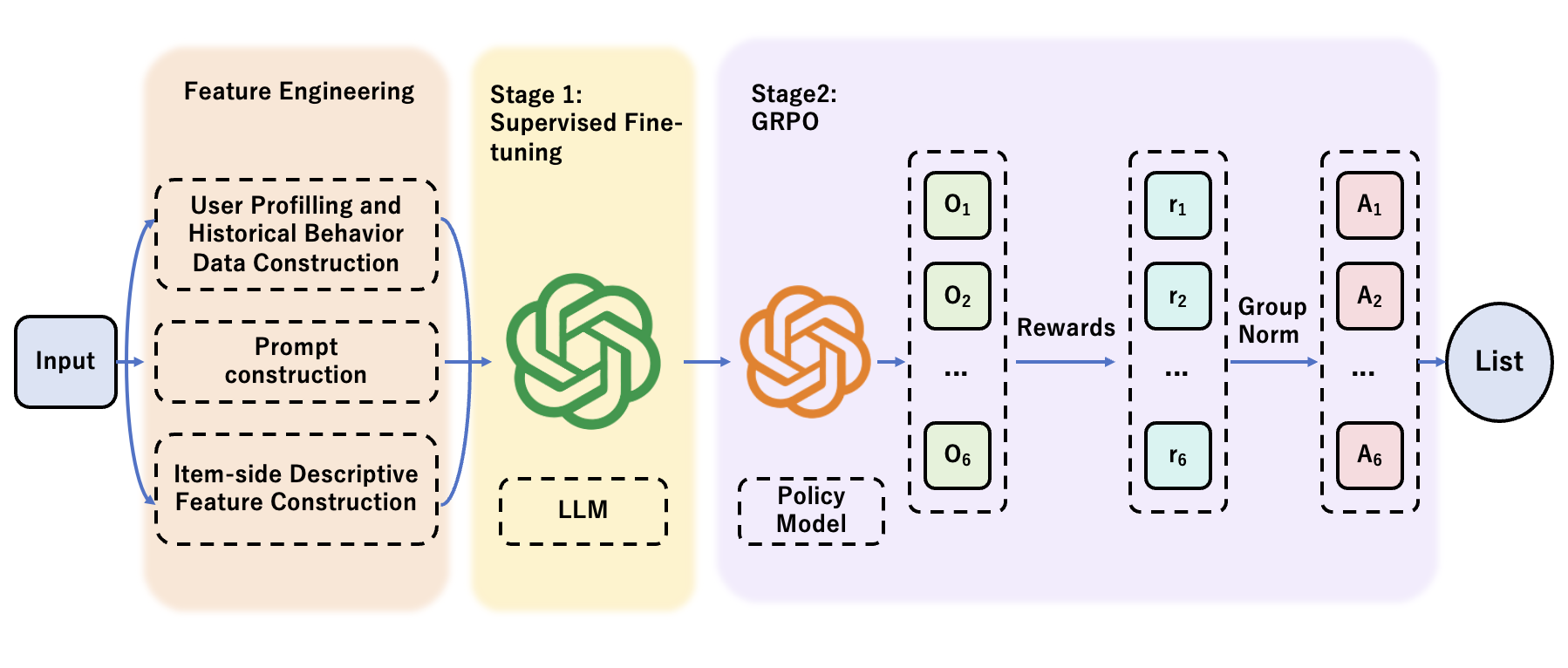}
    \caption{An overview of the modeling approach used in RecLLM-R1.}
    \label{fig: modeling approach}
\end{figure}

\section{Method}
\label{sec:Method}

The proposed recommender system framework comprises three primary components: Data Construction, the SFT Stage, and the GRPO Stage, as depicted in Figure ~\ref{fig: modeling approach}.

\subsection{Data Construction}

To align with the input requirements of large language models, we constructed a training dataset specifically tailored for recommender systems by transforming features into natural language descriptions.
\textbf{User-side Data}:We transformed user profiles (e.g., age, gender, interests) and recent interaction history (e.g., sequences of clicked, viewed, liked, bookmarked, or purchased items) into chronologically ordered textual descriptions, aggregated by topic. For instance, "Over the past 7 days, the user viewed 'Sichuan Cuisine' videos twice and bookmarked one article titled 'Chengdu Travel Guide'."
\textbf{Item-side Data}: We primarily utilized item descriptions, titles, and granular tag labels derived through content understanding, alongside statistical item features, to generate natural language descriptions.
\textbf{Label Construction}: Based on user engagement metrics such as browsing duration and interaction behaviors (e.g., bookmarking, sharing, commenting, following, liking), we constructed a sequence of items that reflects the user's eventual preferences as positive training samples. This ensures the authenticity and business relevance of the labels.
\textbf{Prompt Construction}: We concatenated the user-side and item-side textual information. The objective is to explicitly recommend the most relevant sequence of items to the user, with candidate items drawn from the provided data pool.

Through this methodology, we transformed traditional structured and semi-structured recommendation data into a natural language format interpretable and processable by large language models (LLMs). This data construction approach not only preserves the benefits of traditional interaction data but also imbues large language models with rich semantic context, thereby mitigating the limitations of the "filter bubble" effect.

\subsection{SFT Stage: Activating the Recommender Capability of the Large Language Model}

Following the construction of the textualized recommendation dataset, we proceeded to the Supervised Fine-Tuning (SFT) stage. The primary objective of this phase was not to immediately achieve optimal recommendation performance, but rather to "activate" or "cold-start" the pre-trained large language model (LLM), endowing it with the foundational ability to address recommendation tasks.

For efficiency, we utilized the pre-trained DeepSeek-R1-Distill-Qwen-1.5B model and employed the compact yet high-quality dataset constructed previously. In this stage, the emphasis was on leveraging a minimal amount of data to rapidly unlock the LLM’s potential within the recommendation domain, enabling it to learn the requisite input-output formats and develop a rudimentary understanding of recommendation task modeling. This process prepares the model for the subsequent, more intricate multi-step optimization.

Essentially, the SFT stage serves as a "warm-up" providing the model with baseline recommendation capabilities prior to reinforcement learning. While SFT can already address certain recommendation needs—particularly cold-start scenarios, owing to the LLM’s general knowledge—it is inherently insufficient for optimizing complex long-term objectives, balancing multiple recommendation facets (such as diversity and novelty), or executing advanced reasoning. Therefore, SFT acts as a critical precursor, priming the model for comprehensive optimization in the subsequent GRPO stage. This design closely mirrors the training paradigm of DeepSeek R1.

\subsection{GRPO Stage: Reinforcement Learning and Chain-of-Thought Optimization }
After establishing foundational capabilities in the Supervised Fine-Tuning (SFT) stage, we employ the Group Relative Policy Optimization (GRPO) algorithm ~\cite{shao2024deepseekmath} for reinforcement learning. This phase is dedicated to further enhancing the model's recommendation performance, refining its reasoning capabilities, and ensuring closer alignment with complex, multifaceted business objectives. GRPO, a variant of Proximal Policy Optimization (PPO), optimizes the recommendation policy $\pi_{\theta}$ by maximizing the following objective function:

\begin{align*}
J_{\text{GRPO}}(\theta) =\ & 
\mathbb{E}_{q \sim P_{\text{context}},\ \{o_k\}_{k=1}^G \sim \pi_{\theta_{\text{old}}}(O|q)} \bigg[ \\
& \frac{1}{G} \sum_{k=1}^{G} \frac{1}{|o_k|} \sum_{t=1}^{|o_k|} 
\bigg( \frac{\pi_\theta(o_{k,t} \mid q, o_{k,<t})}
           {\pi_{\theta_{\text{old}}}(o_{k,t} \mid q, o_{k,<t})} \hat{A}_{k,t} \\
& \qquad\qquad
- \beta \left(
    \frac{\pi_{\text{ref}}(o_{k,t} \mid q, o_{k,<t})}
         {\pi_\theta(o_{k,t} \mid q, o_{k,<t})}
    - \log \frac{\pi_{\text{ref}}(o_{k,t} \mid q, o_{k,<t})}
                 {\pi_\theta(o_{k,t} \mid q, o_{k,<t})}
    - 1
\right) \bigg) \bigg]
\end{align*}

In this formulation, $q$ represents the current user context or state (e.g., historical interactions, current query).
 $o_k$ is an output sequence generated by the policy, encompassing both the reasoning steps and the final recommended items. A group of $G$ such sequences is sampled from the old policy $\pi_{\theta_{old}}$.
 $\pi_{\theta}(o_{k,t} | q, o_{k,<t})$ is the probability assigned by the current policy $\pi_{\theta}$ to the $t$-th token $o_{k,t}$ in the $k$-th output sequence, given the context $q$ and preceding tokens $o_{k,<t}$.
 $\hat{A}_{k,t}$ represents the estimated advantage of generating token $o_{k,t}$. Crucially, in GRPO, this advantage is derived from comparing the rewards of the $G$ output sequences within the group, effectively estimating a baseline from these group scores without requiring a separate critic model. This makes the training process more resource-efficient.
 The term scaled by $\beta$ is a KL divergence-related penalty that regularizes the current policy $\pi_{\theta}$ against a reference policy $\pi_{ref}$ (often the SFT model), preventing the model from deviating too drastically and ensuring stability.
This GRPO framework allows the model to learn from experience, iteratively improving its strategy based on the received rewards.

\textbf{Constructing the Reinforcement Learning Framework.} The recommendation task is framed as a multi-step decision process. We specifically employ the GRPO algorithm, a policy gradient method, to train the model. This approach ensures that the model learns not just to optimize for immediate user responses but also to enhance long-term user experience and achieve sustained engagement.

\textbf{Introducing Chain-of-Thought (CoT).} To bolster the model's reasoning capabilities, we integrate a Chain-of-Thought (CoT) mechanism into each recommendation decision. Before outputting the final recommended sequence (which forms part of $o_k$ in the GRPO objective), the model first generates the underlying reasoning for its choices. This multi-step reasoning process, manifest as intermediate tokens in $o_k$, guides the model towards more interpretable and contextually appropriate recommendations, and the entire generation (CoT + recommendation) is evaluated for the reward signal.

\textbf{Reward Function Design.} The design of a flexible and comprehensive reward function $R$ is pivotal to the success of GRPO, as the signal from $R$ is used to compute the advantage estimates $\hat{A}_{k,t}$ in the GRPO objective function. Our reward function $R$ is a composite function that evaluates multiple facets of the recommendation output $o_k$. It can combine traditional recommendation metrics (e.g., click-through rate, conversion rate) with business-specific customized metrics (e.g., diversity of recommended items, proportion of new or emerging content, user retention indicators). This carefully designed reward structure enables the GRPO framework to achieve tailored optimization for diverse and evolving business scenarios. This approach not only improves the strategic alignment and rationality of the model’s decisions but also offers the necessary flexibility for robust industrial applications, ensuring the reward function remains simple and clear in its objectives.

By first activating basic capabilities through SFT and then employing GRPO combined with CoT for reinforcement learning optimization, our framework aims to train a robust recommendation model. This model leverages the rich knowledge and reasoning abilities inherent in Large Language Models and dynamically adapts to user feedback and complex optimization goals via RL. It is thus capable of optimizing its recommendation strategies from both short-term and long-term perspectives, thereby overcoming the limitations of traditional direct iteration training methods that often struggle with long-sequence dependencies and multi-faceted objectives.


\section{Experiments}
\label{sec: Experiments}

This chapter aims to validate the efficacy of our proposed RecLLM-R1 recommendation framework—grounded in the DeepSeek R1 methodology—through a series of experiments conducted on both public benchmarks and a real-world industrial dataset. We will detail the datasets employed, evaluation metrics utilized, implementation specifics, and provide a comparative analysis of our approach against mainstream baseline methods. Finally, an ablation study will be presented to analyze the contribution of key components within the framework.

\subsection{Datasets}

Public Datasets: We evaluated the proposed framework on three publicly available real-world benchmarks from the Amazon Product Reviews dataset ~\cite{he2016ups}, which encompasses user reviews and item metadata spanning from May 1996 to July 2014. Specifically, we utilized three categories of this dataset for the sequential recommendation task: “Sports and Outdoors,” “Beauty,” and “Toys and Games.”

Real-World Industrial Dataset: The experimental data originates from anonymized real user behavior logs on a large-scale social media platform (currently under review for public release), containing comprehensive user profiles, historical interaction records, and detailed item descriptions, titles, and labels. This dataset primarily comprises three types of information:
\textbf{User Data}: This includes unique user identifiers and static profile attributes such as age, gender, city tier, and a set of long-term interest tags.
\textbf{Item Data}: This encompasses unique item identifiers, textual content (titles and body descriptions), fine-grained tags extracted via content understanding models, and item-level statistics (e.g., number of likes, comments, shares).
\textbf{Interaction Data}: Includes users’ short- and long-term preferences across different content types (e.g., counts of clicks, likes, saves, comments, shares).
\textbf{Data Preprocessing:}: To mitigate noise from cold-start users, we removed users with fewer than five interactions. Textual data underwent a cleaning process, including the removal of special characters and format unification.
\textbf{Sequence Construction}: To prevent data leakage, each user’s interaction history was strictly ordered by timestamp. Due to LLM context window limitations and potential information decay, we segmented user histories into short-term and long-term sequences and converted them into natural language descriptions.
\textbf{Label Construction}: Positive item sequences reflecting users’ genuine preferences were constructed based on strong engagement signals such as prolonged dwell time and deep interactions (likes, saves, comments, shares, follows).
\textbf{Evaluation Metrics}: We employed top-k Recall (Recall@K) and Normalized Discounted Cumulative Gain (NDCG@K) with K = 5 and 10 to evaluate the recommendation performance.

\subsection{Implementation Details}

For fine-tuning, we utilized DeepSeek-R1-Distill-Qwen-1.5B as our base large language model. This model, characterized by a relatively smaller parameter count yet strong performance, is well-suited for efficient iteration under constrained computational resources. All experiments were conducted on eight NVIDIA H800 GPUs, each with 80GB of memory. The implementation was based on the VeRL library. User profiles, historical interaction sequences, and candidate item sets were transformed into natural language prompts.

SFT stage: To activate the LLM’s capability for sequential recommendation, we performed supervised fine-tuning using a compact, high-quality subset of the training data.

GRPO stage: Building upon the model obtained from the SFT stage, we initialized a policy network that was conditioned to first generate one or more "reasoning steps" (i.e., recommendation rationales) before producing the final item ID sequence. For each GRPO iteration, we generated twelve candidate recommendation lists per prompt. These lists were scored using the reward function, and the group-relative loss was computed to update the policy network. Our model was trained using the AdamW optimizer with an initial learning rate of $1 \times 10^{-4}$.

Reward Function: The core of our reward mechanism is a normalized, position-weighted Longest Common Subsequence (LCS) algorithm, where higher rewards are assigned to predictions with more correct items ranked at earlier positions. To mitigate the recurrence of common errors during training, we incorporated specific penalties. All other reward components adopted the default formats and reasoning-based reward functions provided by VeRL\cite{sheng2024hybridflow}.

\subsection{Recommendation Performance}

\begin{table}[ht]
\centering
\caption{Performance comparison of different methods on public datasets.}
\resizebox{\textwidth}{!}{
\begin{tabular}{lccccccc}
\toprule
\textbf{Metric} & GRU4Rec & BERT4Rec & FDSA & SASRec & S$^3$-Rec & TIGER & \textbf{recLLM-R1 (Ours)} \\
\midrule
\multicolumn{8}{c}{\textit{Sports and Outdoors}} \\
Recall@5 & 0.0129 & 0.0115 & 0.0182 & 0.0233 & 0.0251 & \uline{0.0264} & \textbf{0.0322 (+21.97\%)} \\
NDCG@5   & 0.0086 & 0.0075 & 0.0122 & 0.0154 & 0.0161 & \uline{0.0181} & \textbf{0.0237 (+30.94\%)} \\
Recall@10 & 0.0204 & 0.0191 & 0.0288 & 0.0350 & 0.0385 & \uline{0.0400} & \textbf{0.0471 (+17.75\%)} \\
NDCG@10 & 0.0110 & 0.0099 & 0.0156 & 0.0192 & 0.0204 & \uline{0.0225} & \textbf{0.0302 (+34.22\%)} \\
\midrule
\multicolumn{8}{c}{\textit{Beauty}} \\
Recall@5 & 0.0164 & 0.0203 & 0.0267 & 0.0387 & 0.0387 & \uline{0.0454} & \textbf{0.0541 (+19.16\%)} \\
NDCG@5   & 0.0099 & 0.0124 & 0.0163 & 0.0249 & 0.0244 & \uline{0.0321} & \textbf{0.0405 (+26.17\%)} \\
Recall@10 & 0.0283 & 0.0347 & 0.0407 & 0.0605 & 0.0647 & \uline{0.0648} & \textbf{0.0729 (+12.50\%)} \\
NDCG@10 & 0.0137 & 0.0170 & 0.0208 & 0.0318 & 0.0327 & \uline{0.0384} & \textbf{0.0459 (+19.53\%)} \\
\midrule
\multicolumn{8}{c}{\textit{Toys and Games}} \\
Recall@5 & 0.0097 & 0.0116 & 0.0228 & 0.0463 & 0.0443 & \uline{0.0521} & \textbf{0.0596 (+14.40\%)} \\
NDCG@5   & 0.0059 & 0.0071 & 0.0140 & 0.0306 & 0.0294 & \uline{0.0371} & \textbf{0.0411 (+10.78\%)} \\
Recall@10 & 0.0176 & 0.0203 & 0.0381 & 0.0675 & 0.0700 & \uline{0.0712} & \textbf{0.0773 (+8.57\%)} \\
NDCG@10 & 0.0084 & 0.0099 & 0.0189 & 0.0374 & 0.0376 & \uline{0.0432} & \textbf{0.0508 (+17.60\%)} \\
\bottomrule
\end{tabular}
}
\label{tab:public_compare}
\end{table}

\begin{table}[ht]
\centering
\caption{Performance comparison of different methods on Industrial datasets.}
\begin{tabular}{l|cc|cc}
\toprule
Method & Recall@5 & NDCG@5  & Recall@10 & NDCG@10  \\
\midrule
online                & 0.3381 & 0.3762 & 0.4053 & 0.4802  \\
\textbf{recLLM-R1[Ours]}      & \textbf{0.4137} & \textbf{0.4692}  & \textbf{0.5311} & \textbf{0.5653}  \\
\bottomrule
\end{tabular}
\label{tab:online}
\end{table}

Performance on Public Datasets (Table ~\ref{tab:public_compare}): The experimental results unequivocally demonstrate that RecLLM-R1 attains the highest performance across all four evaluation metrics (Recall@5, NDCG@5, Recall@10, and NDCG@10) on all three public datasets (Sports and Outdoors, Beauty, Toys and Games), in comparison to all baseline models, including GRU4Rec ~\cite{hidasi2015session}, BERT4Rec ~\cite{sun2019bert4rec}, FDSA ~\cite{zhang2019feature}, SASRec ~\cite{kang2018self}, S$^3$-Rec ~\cite{zhou2020s3}, and TIGER ~\cite{rajput2023recommender}. Notably, when contrasted with the best-performing baseline, RecLLM-R1 exhibits substantial improvements across various datasets and metrics. For instance, it achieved a 34.22\% increase in NDCG@10 on the Sports and Outdoors dataset, a 26.17\% increase in NDCG@5 on the Beauty dataset, and a 17.60\% increase in NDCG@10 on the Toys and Games dataset. These findings consistently underscore the superior performance of our method over current state-of-the-art sequential recommendation models on established public benchmarks across diverse domains.

Performance on Industrial Dataset (Table ~\ref{tab:online}): On the real-world industrial dataset, our RecLLM-R1 method also demonstrates exceptional performance. Compared to the online baseline system, RecLLM-R1 yielded significant gains across Recall@5, NDCG@5, Recall@10, and NDCG@10 metrics. For example, Recall@10 improved from 0.4053 to 0.5311, and NDCG@10 increased from 0.4802 to 0.5653. This robustly validates the effectiveness and superiority of our method in practical industrial application scenarios.

In summary, comprehensive experiments conducted on both public and industrial datasets conclusively demonstrate that our proposed RecLLM-R1 method achieves robust performance and strong generalization in sequential recommendation tasks. RecLLM-R1 consistently and significantly outperforms existing baselines and state-of-the-art models in both standard benchmarks and real-world industrial settings. These results underscore the considerable potential of synergistically combining advanced semantic understanding from large language models with Chain-of-Thought-based reinforcement learning within sequential recommendation frameworks. The substantial and consistent improvements observed indicate that this approach more effectively captures subtle nuances in user preferences and item characteristics, leading to enhanced prediction accuracy. Furthermore, by leveraging flexible reward function design, our method exhibits an improved capacity to incorporate intricate business strategies relevant to real-world applications. Overall, this work highlights a promising trajectory for the future evolution of recommendation systems, emphasizing the increasing importance of deep semantic understanding and logical reasoning in advancing recommender system capabilities.

\section{Conclusion}
\label{sec: Conclusion}

This study introduces RecLLM-R1, a novel recommendation framework that integrates Large Language Models (LLMs) with reinforcement learning, drawing inspiration from the DeepSeek R1 methodology. We initially transform user profiles, behavior histories, and item metadata into natural language prompts interpretable by LLMs. Subsequently, a two-stage training strategy is employed: (1) Supervised Fine-Tuning (SFT) to activate the model’s recommendation capabilities using a focused set of high-quality data; and (2) Group Relative Policy Optimization (GRPO), enhanced with Chain-of-Thought (CoT) reasoning, to directly optimize multi-objective business metrics through reinforcement learning. Empirical evaluations on public and real-world user behavior datasets demonstrate that RecLLM-R1 significantly surpasses traditional recommenders and basic LLM methods in accuracy, diversity, and novelty, offering a robust solution to address the limitations inherent in conventional recommendation systems.

\bibliographystyle{plain}
\bibliography{ref.bib}

\end{document}